\pdfoutput=1
\documentclass[sigconf]{acmart}
\copyrightyear{2023}
\acmYear{2023}
\setcopyright{rightsretained}
\acmConference[CIKM '23]{Proceedings of the 32nd ACM International Conference on Information and Knowledge Management}{October 21--25, 2023}{Birmingham, United Kingdom}
\acmBooktitle{Proceedings of the 32nd ACM International Conference on Information and Knowledge Management (CIKM '23), October 21--25, 2023, Birmingham, United Kingdom}\acmDOI{10.1145/3583780.3615109}
\acmISBN{979-8-4007-0124-5/23/10}
\usepackage{subfigure}
\usepackage{graphicx}
\usepackage{booktabs}
\usepackage{url}
\usepackage{tabularx}
\usepackage{bigstrut}
\usepackage{multirow}
\usepackage{color}
\usepackage{xcolor}
\usepackage{balance} 
\usepackage{verbatimbox}

\usepackage{enumitem}
\usepackage{xspace}
\usepackage{utfsym}
\usepackage{algpseudocode}
\usepackage{algorithm}

\newcommand{\our}{WOT-Class\xspace}
\newcommand{\smallsection}[1]{\noindent\textbf{#1.}}

\makeatletter
\gdef\@copyrightpermission{
  \begin{minipage}{0.3\columnwidth}
   \href{https://creativecommons.org/licenses/by/4.0/}{\includegraphics[width=0.90\textwidth]{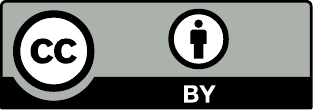}}
  \end{minipage}\hfill
  \begin{minipage}{0.7\columnwidth}
   \href{https://creativecommons.org/licenses/by/4.0/}{This work is licensed under a Creative Commons Attribution International 4.0 License.}
  \end{minipage}
  \vspace{5pt}
}
\makeatother
\begin{document}

\title{\our: Weakly Supervised Open-world Text Classification}

\author{Tianle Wang}
\email{wtl666wtl@sjtu.edu.cn}
\affiliation{
\institution{Shanghai Jiao Tong University}
\city{Shanghai}
\country{China}
}
\author{Zihan Wang}
\email{ziw224@ucsd.edu}
\affiliation{
\institution{University of California, San Diego}
\city{La Jolla}
\state{CA}
\country{USA}
}
\author{Weitang Liu}
\email{wel022@ucsd.edu}
\affiliation{
\institution{University of California, San Diego}
\city{La Jolla}
\state{CA}
\country{USA}
}
\author{Jingbo Shang}
\email{jshang@ucsd.edu}
\authornote{Corresponding Author.}
\affiliation{
\institution{University of California, San Diego}
\city{La Jolla}
\state{CA}
\country{USA}
}

\renewcommand{\shortauthors}{Tianle Wang, Zihan Wang, Weitang Liu, \& Jingbo Shang}

\begin{abstract}
State-of-the-art weakly supervised text classification methods, while significantly reduced the required human supervision, still requires the supervision to cover all the classes of interest.
This is never easy to meet in practice when humans explore new, large corpora without complete pictures.
In this paper, we work on a novel yet important problem of weakly supervised open-world text classification, where supervision is only needed for \emph{a few examples from a few known classes} and the machine should handle both \emph{known} and \emph{unknown classes} in test time. 
General open-world classification has been studied mostly using image classification; however, existing methods typically assume the availability of sufficient known-class supervision and strong unknown-class prior knowledge (e.g., the number and/or data distribution).
We propose a novel framework \our that lifts those strong assumptions. 
Specifically, it follows an iterative process of (a) clustering text to new classes, (b) mining and ranking indicative words for each class, and (c) merging redundant classes by using the overlapped indicative words as a bridge.
Extensive experiments on 7 popular text classification datasets demonstrate that \our outperforms strong baselines consistently with a large margin, attaining 23.33\% greater average absolute macro-F1 over existing approaches across all datasets.
Such competent accuracy illuminates the practical potential of further reducing human effort for text classification.

\end{abstract}




\maketitle

\section{Introduction}

Weakly supervised text classification methods, including zero-shot prompting, can build competent classifiers from raw texts by only asking humans to provide (1) a few examples per class~\cite{DBLP:conf/cikm/MengSZ018,DBLP:conf/acl/GaoFC20} or (2) class names~\cite{DBLP:journals/corr/abs-2005-14165,DBLP:conf/emnlp/MengZHXJZH20, wang-etal-2021-x}. 
All these methods require that the human-provided \emph{known classes} cover all the classes of interest, however, it can be very difficult especially in the dynamic and ever-changing real world.
For example, the human expert could be exploring a new, large corpus without a complete picture.

\begin{figure}[t!]
\begin{center}
\includegraphics[width=\linewidth]{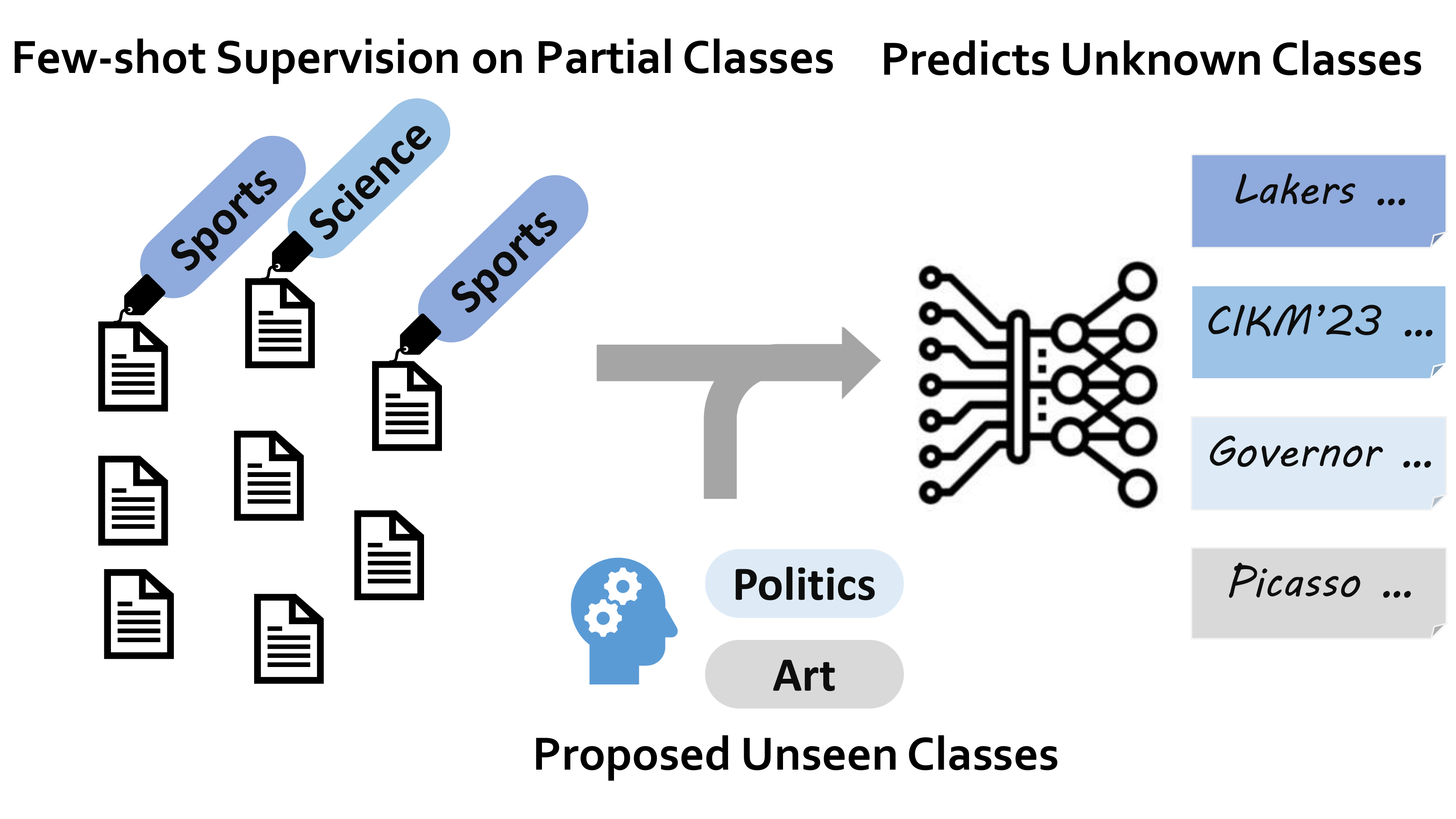} 
\caption{Weakly Supervised Open-World Text Classification. We aim to cluster text in a corpus, where only a few classes have few-shot supervision and class names known.
}
\label{fig:intro}
\end{center}
\end{figure}

In this paper, we work on a novel yet important problem of weakly supervised open-world text classification as shown in Figure~\ref{fig:intro}.
Specifically, the human is only asked to provide a few examples for every known class; 
the machine is tasked to dive into the raw texts, discover possible \emph{unknown classes}, and classify all the raw texts into corresponding classes, including both known and unknown.
The open-world setting here releases the all-class requirement, further reducing the required human effort in weakly supervised text classification.
We argue that this problem is feasible because one could expect that unknown classes follow a similar taste as the known classes, i.e., the classes should follow certain underlying high-level semantic meanings and the same granularity level. 
For example, if the known classes are ``\emph{Awesome}'' and ``\emph{Good}'', one would expect to see classes like ``\emph{Terrible}'' and ``\emph{Bad}'';
in Figure~\ref{fig:intro}, ``\emph{Politics}'' can be a reasonable choice for unknown class.

\begin{figure*}[t!]
\begin{center}
\includegraphics[width=\textwidth]{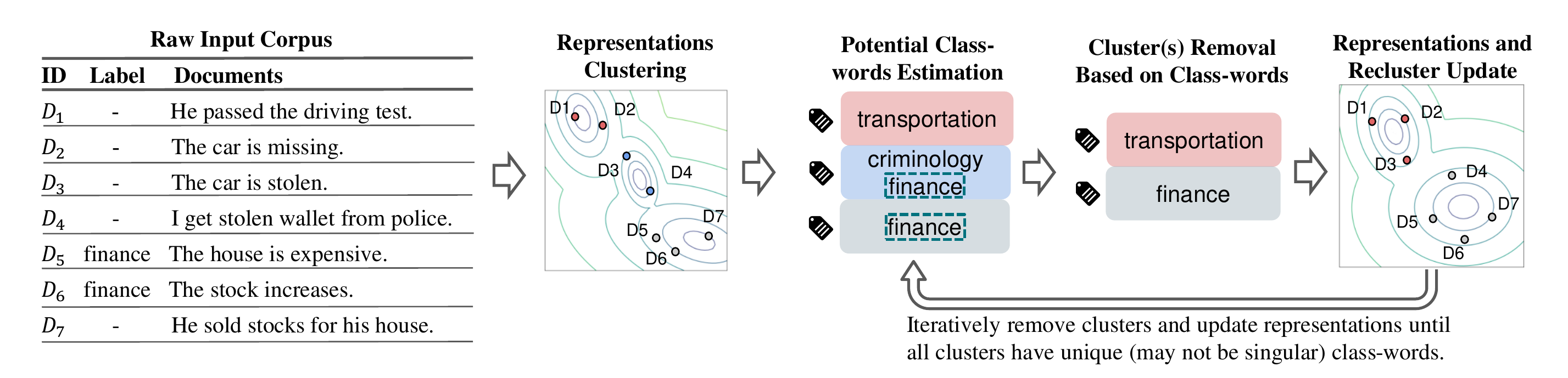} 
\caption{An overview of \our framework.
Given the corpus and a part of labels, we first estimate document representations and construct the initial clusters. And then, we perform an iterative cluster refinement to remove redundant clusters. At the end of each iteration, we will update the document representations and recluster them.
}
\label{fig:pipeline}
\end{center}
\end{figure*}

Open-world classification~\cite{shu2018unseen, xu2019open, han2021autonovel, cao2021openworld, vaze2022gcd} 
has been studied, mostly in image classification; however, existing methods typically assume the availability of sufficient known-class supervision and strong unknown-class prior knowledge (e.g., the number and/or data distribution).
Text classification has its uniqueness --- The text is composed of words, some of which reflect the semantics of the classes.
These class-indicative words (abbreviated as \emph{class-words}) are the key source of weak supervision signals~\cite{tao2015doc2cube,mekala-shang-2020-contextualized,wang-etal-2021-x}.
As shown in Figure~\ref{fig:intro}, class-words such as ``\emph{Governor}'' can help discover the unknown class \emph{Politics} and its related documents.
Such a feature motivates us to design a new open-world framework particularly for weakly supervised text classification. 

We propose a novel, practical framework \our\footnote{\our means \emph{What class? Let's discover!}}, which lifts those strong assumptions of existing methods.
Figure~\ref{fig:pipeline} illustrates the general idea of \our. 
It leverages the class-words in text to iteratively refine the text clustering and ranking of class-words.
Specifically, we first make an overestimation of the number of classes and construct initial clusters of documents based on the names of known classes.
Then, we employ an iterative process to refine these clusters.
We first select a set of candidate class-words for them through statistics and semantics.
Then we learn a classifier to provide each cluster a ranking of class-words based on the limited known-class supervision. 
When there is redundancy among these clusters, the high-ranked class-words for the clusters will overlap, in which case we know at least one cluster is redundant.
The refined set of class-words will help re-cluster documents, and we repeat this process till the number of classes no longer decreases.

We conduct our experiments by fixing the most infrequent half of classes as unseen, which emphasizes the imbalanced and emerging nature of real-world scenarios.
And our extensive experiments on 7 popular text classification datasets have shown the strong performance of \our. 
By leveraging merely a few known-class examples and the names of known classes, \our gains a 23.33\% greater average absolute macro-F$_1$ over the current best method across all datasets. When given 
our prediction of classes as an extra input, \our still achieves 21.53\% higher average absolute macro-F$_1$.
While precisely discovering unseen classes identical to the ground truth remains challenging, our method can provide predictions closest to the actual classes more stably than existing approaches. 
And considering \our provides class-words for each discovered unknown class, it shall only require a reasonable amount of effort for humans to digest the discovered ones to something similar to the ground truth. 
Finally, \our is less sensitive to class imbalances, making it more suitable in real-world scenarios.

Our contributions are as follows.
\begin{itemize}[leftmargin=*,nosep]
  \item We introduce the novel yet important problem of weakly supervised open-world text classification.
  \item We propose a novel, practical framework \our that jointly performs document clustering and class-word ranking that can discover and merge unknown classes.
  \item Extensive experiments demonstrate that \our outperforms the previous methods in various manners. 
  Competent accuracy of \our also illuminates the practical potential of further reducing human effort for text classification.
\end{itemize} 

\noindent\textbf{Reproducibility.} We release the code and datasets on Github\footnote{\url{https://github.com/wtl666wtl/WOT-Class}}.

\section{Preliminaries}
In this section, we formally define the problem of weakly supervised open-world text classification.
And then, we brief on some preliminaries about CGExpan and X-Class, two building blocks  
that we will use in our method.

\smallsection{Problem Formulation}
In an open-world setting, there exists a not-fully-known set of classes $\mathcal{C}$, which follows the same hyper-concept and a set of documents $\mathcal{D}$, each  
uniquely assigned to a class. 
A weakly supervised open-world model can observe partial information of $\mathcal{C}$. In this work, we assume that partial information is given as a labeled few-shot dataset $\mathcal{D}_s = \{x_i, y_i\}_{i=1}^n, y_i \in \mathcal{C}_s,$ where $\mathcal{C}_s \subset \mathcal{C}$ where $\mathcal{C}_s$ is the known subset of classes and $n$ is rather small (e.g., a ten-shot dataset would mean $n = 10 * |\mathcal{C}_s|$). The goal of the model is to classify the remainder of the dataset, $\mathcal{D}_u = \mathcal{D} \backslash \mathcal{D}_s$, where some of the labels in $\mathcal{C}_u = \mathcal{C} \backslash \mathcal{C}_s$ is completely unknown to the model. 
We emphasize that different from extremely weakly supervised or zero-shot prompting based text classifiers, the names of the unknown classes are also not given to the model.

\smallsection{CGExpan}\label{sec:cgexpan}
Entity set expansion aims to expand a set of seed keywords (e.g., \textit{United Sates}, \textit{China}) to new keywords (e.g., \textit{Japan}) 
following the same hyper-concept (i.e., \textit{Country}).
This is the exact technique to help us discover potential class words that are highly suggestive of the hidden class names.
In our method, we employ CGExpan~\cite{zhang-etal-2020-empower}, one of the current state-of-the-art methods for set expansion.
CGExpan selects automatically generated hyper-concept words by probing a pre-trained language model (e.g., BERT), and further ranks all possible words guided by selected hyper-concept.
However, a common problem of such set expansion method is that they typically give duplicated and semantically-shifted entities even at the top of the rank list.
In our work, we utilize CGExpan to find semantically related words to the user-given class names as \textit{candidates} for the class-words. 
Our method resolves this imperfect set of candidates problem by ranking them based on a scoring metric learned by few-shot supervision.

\smallsection{X-Class}\label{sec:xclass}
X-Class is an extremely weakly supervised text classification method that works with only names of the classes~\cite{wang-etal-2021-x}. It proposes a framework that learns representations of classes and text, and utilizes clustering methods, such as a Gaussian Mixture Model~\cite{reynolds2009gaussian} to assign text to classes.
While X-Class showed promising performance in close-world classification settings with minimal supervision, it does not work in open-world settings. 
Our method for the open-world classification problem reduces open-world text classification to close-world classification by iterative refinement of class-words. 
Therefore, a strong performing (and efficient) close-world text classifier X-Class is employed.

\smallsection{Static Representation}
For each unique word $w$ in the input corpus, we obtain its \textit{static representation} $\mathbf s_w$ by averaging BERT's contextualized representations of all its appearances. That is,

\begin{equation*}
\mathbf s_w = \frac{\sum_{w^\prime = w} \mathbf t_{w^\prime}}{\sum_{w^\prime = w} 1},
\end{equation*}
where $w^\prime$ are occurrences of the word in the corpus and $\mathbf t_{w^\prime}$ is its contextualized word representation\footnote{For a word that can be split into multiple tokens, its contextualized word representation is obtained by taking the average of the contextualized representations of all its constituent tokens.}. A static representation is useful in determining the similarity and relatedness of two words.
\begin{figure*}[h!]
\begin{center}
\includegraphics[width=0.92\textwidth]{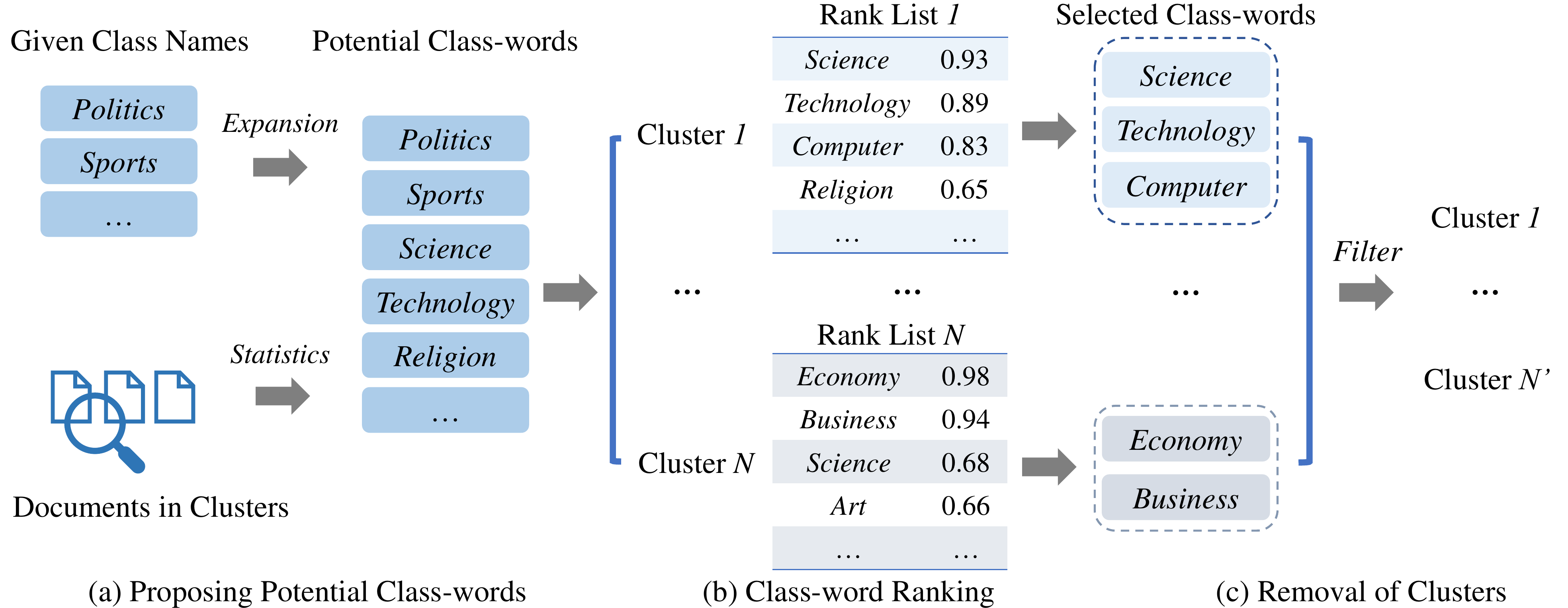} 
\caption{An overview of \emph{Cluster} $\to$ \emph{Class-words}.
}
\label{fig:method}
\end{center}
\end{figure*}
\section{Our \our Method}
In this section, we introduce our \our framework.
To be able to accompany unknown classes, a common approach~\cite{cao2021openworld} is to first overestimate the number of classes and then reduce them after observing the data. 
We follow this approach and integrate it with class-words, with the goal of reducing the problem into an existing weakly supervised text classification (WS-TC) problem, where there are solutions to classify text in a close-world setting when class-words are given for each class and all classes are known.
In simple words, \our first proposes a set of high-potential words from which class-words can be found, and a list of class-words for an over-estimated number of classes. 
At each iteration, we start with clusters of documents obtained by WS-TC methods on the proposed class-words, and rerank the high-potential words to find the new list of class-words for each class. 
During this, classes of similar class-words are removed, and a new cluster of documents can be obtained by WS-TC methods.
The iteration stops when no class removal happens.
We summarize our iterative refinement framework in Algorithm~\ref{1}.
And Figure~\ref{fig:method} shows an overview of the sub-components of this method.

\subsection{Initial Overestimation and Clustering}
In \our, we take the approach where we make an initial overestimation of classes, and then try to refine the class-words and clusters of text. 
We fix a number $K$ ($K = 100$ in all experiments) and ask CGExpan (refer to Section~\ref{sec:cgexpan}) to suggest $K - |\mathcal{C}_s|$ similar words as the given $|\mathcal{C}_s|$ class names. We consider these as a rough approximation of classes in the corpus, and employee X-Class (refer to Section~\ref{sec:xclass}) to construct the initial clusters of text, that may contain many duplicates and different granularity clusters. 
From now on, our iterative method is completed by two processes. In the first process, we obtain a list of class-words for each cluster and remove duplicate ones; in the second process, we simply apply X-Class to refine the clusters based on the new class-words.
We mainly elaborate on the first process.

\begin{algorithm}[t!]\normalsize
		\caption{\our's Iterative Refinement Framework}\label{1}
		\begin{algorithmic}[1]
        	\Require clusters $\mathbf{C}$, document representations $\mathbf{R}$ 
            \While{\text{there are still redundant clusters}}
            \State Find class-indicative words $\mathbf{W}$
            \For{\text{each cluster $\mathbf{C}_i$ in $\mathbf{C}$}}
            \State Train MLP and rank $\mathbf W$
            \State Select possible names $\mathbf{S}_i$ from $\mathbf{W}$
            \State Compute cluster 
            coherence $ \eta_i$ 
            (Eq. \ref{equ:purity_score})
            \EndFor
            \For{\text{each pair $\mathbf{S}_i, \mathbf{S}_j $}}
           	\If{$\mathbf{S}_i\cap \mathbf{S}_j \neq \emptyset$ and $\eta_i \leq \eta_j$ }
           	\State Remove $\mathbf{C}_i$
           	\EndIf
            \EndFor
            \State Re-estimate class names $\mathbf{S}$
            \State update $\mathbf{R}, \mathbf{
            C}$ based on $\mathbf S$
            \EndWhile
		\end{algorithmic}
	\end{algorithm}

\subsection{Cluster $\to$ Class-words}\label{sec:class-word}
In the first process, we start with clusters of text, the initially suggested words by CGExpan, the class-words in the last iteration (for the first iteration, the class-words are the CGExpan proposed words) and the few-shot supervision, and aim to reduce the number of clusters and assign class-words to each cluster.

\smallsection{Proposing Potential Class-words} 
Class-words are words that are related to and highly indicative of the class. Words from CGExpan qualify for the relativeness, but we also wish to find words that are rather exclusive to all except one (or a few, because of the noise in the clusters) cluster. 
The indicativeness of a word can be expressed by the statistical outstandness of it to its cluster of text, compared to other clusters. Such statistical measures are well-researched in information retrieval, the representative would be the tf-idf~\cite{ramos2003using} score. We apply a more recent measure that has been used in text classification~\cite{mekala-shang-2020-contextualized} to find statistically representative words within cluster $i$:
\begin{equation*}
score_i(w)=\frac{s_i(w)}{size_i}\cdot\tanh\left(\frac{t_i(w)}{size_i}\right)\cdot\log\left(\frac{|\mathcal D|}{s_{\mathcal D}(w)}\right),
\end{equation*}
where $t_i(w)$ is the number of occurrences of the word $w$ in documents belonging to cluster $i$, $s_i(w)$ indicates how many documents in cluster $i$ contain the word $w$ while $s_{\mathcal D}(w)$ indicates the document frequency of word $w$, and $size_i$ indicates how many documents are in cluster $i$.
In the measurement, the first term tells how indicative a word is to a cluster, the second term measures how frequent this word is, and the third is a normalization based on the inverse document frequency.

We find the top such statistically representative words for each cluster and merge these statistically representative words with words from CGExpan as the set of potential class-words.

\smallsection{Class-word Ranking}
We utilized CGExpan and statistical representativeness to approximately retrieve a list of potential class-words, and now we precisely rank them by learning a metric that defines the similarity between a cluster and a word.

Specifically, 
we construct features for a potential class-word to a cluster as the mean and variance of Euclidean distance and cosine similarity of static representations between 
the class-word and (a large list of $W = 50$) statistically representative words for the cluster\footnote{A larger list of statistically representative words is further employed for each cluster to detect the closeness with potential class-words. When proposing class-words, note we only select the top statistically representative words in each cluster.}. Since we know a few labeled examples, they serve as virtual clusters where we treat the features of their class names to the respective virtual clusters as positive signals of closeness. The negative signals are derived through an intuitive heuristic where we find the most dissimilar word (i.e., the word with the furthest static representation from the known class name) from the set of potential class-words. With these two signals, we train a Multilayer Perceptron (MLP) logistic regression on the features to predict the signal. 
We assign a score $p(w, i)$ for each cluster $i$ and each word from the high potential words. 

We also propose a post-processing step to remove generic words from the ranking. We follow previous work~\cite{Jones72astatistical} and design a penalty coefficient $\mu(w, i)$ for each candidate class name $w$ in cluster $i$ based on inter-class statistics:
\begin{equation*}
\mu(w,i)=\log \left(\frac{\textsc M  \{rank_j(w)\mid 1\leq j \leq C \}}{1+rank_i(w)}\right),
\end{equation*}
where $rank_i(w)$ is the absolute rank number of $w$ in cluster $i$ based on MLP's prediction, $\textsc M\{\cdot\}$ indicates the median value of the set, and $C$ is the size of the clusters in the current iteration. 

The main idea of this formula is to obtain a coefficient to penalize those generic words (e.g., life, which might rank high in most clusters) from being selected as class-words. The numerator of the fraction shows how the word behaves across all clusters while the denominator shows how it behaves in a specific cluster. The median rank of a generic word will be very close to the specific rank. Note that we allow one word as the class name of several clusters because of the initial overestimation, but if a word ranks high in more than half of the clusters, it is considered a generic word that must be filtered.
Such penalization and normalization are similar to the idf term in information retrieval. Therefore, we follow the design and choose to divide the two values and take the logarithm. Similar to the idf, this penalty coefficient lowers the chance of selecting a generic word but will not harm proper words.

The final indicativeness ranking is based on the product of two scores:
\begin{equation*}
I(w, i) = p(w, i) \times \mu(w, i).
\end{equation*}

\begin{table*}[t!]
  \begin{center}
  \scriptsize
    \caption{An overview of our datasets. The imbalance factor refers to the ratio of sample sizes between the most frequent class and least frequent one in the dataset.\label{tab:0}}
    \resizebox{.95\linewidth}{!}{
    \begin{tabular}{c  c  c  c  c  c  c  c}
      \toprule
        & \textbf{AGNews} & \textbf{20News} & \textbf{NYT-Small} & \textbf{NYT-Topics} & \textbf{NYT-Locations} & \textbf{Yahoo} & \textbf{DBpedia}\\
      \midrule
      Corpus Domain & News & News & News & News & News & QA & Wikipedia\\
      Class Criterion & Topics & Topics & Topics & Topics & Locations & Topics & Ontology\\
      \# of Classes & 4 & 5 & 5 & 9 & 10 & 10 & 14 \\
      \# of Documents & 12,000 & 17,871 & 13,081 & 31,997 & 31,997 & 18,000 & 22,400 \\
      Imbalance & 1.00 & 2.02 & 16.65 & 27.09 & 15.84 &  1.00 & 1.00\\
      \bottomrule
    \end{tabular}}
  \end{center}
\end{table*}

\smallsection{Removal of Clusters}
We finally discuss how we remove the clusters based on the class-words found. In simple terms, we remove clusters that have non-empty intersections in the class-words. While the cluster is removed, 
all its documents are discarded until they are reassigned to new clusters in the second text clustering process.

Precisely, we pick the $T$ highest ranked class-words for a cluster to compare, where $T$ is the number of iterations in the removal process, a simple way to inject the prior that the cluster quality is better and better after each iteration. We noticed that in certain clusters, there might not be enough good class-words, so we introduce a cutoff threshold $\beta$ such that we do not pick words that have a low ratio of indicativeness score to the highest indicativeness score in the cluster.
Then, when two list of class words overlap, we would like to retain one cluster (or in other words, the list of class-words) and remove the other. We remove the cluster with a low coherent score $\eta$, which is the closeness of the text in the cluster. This can be obtained from X-Class which provides representations of each text, and we can simply compute
\begin{equation}\label{equ:purity_score}
\eta = \frac{1}{|\mathbf{R}|}\sum_{\mathbf r \in \mathbf{R}} \cos\left(\mathbf r,
\overline{\mathbf{R}}\right),
\end{equation}
where $\mathbf{R}$ is the list of text representations belonging to the cluster. When overlap happens, we remove the cluster that has lower coherence $\eta$.

We also rerank the class words after removing the duplicated clusters and continue until no clusters require removal.

\subsection{Iterative Framework and Final Classifier}
The  whole iterative framework is simply applying the first class-word suggesting and cluster deduplicating process and the second class-word-based text clustering process iteratively, until the first process no longer removes clusters.

After exiting the iterative loop and obtaining the stable set of clusters, we follow the common approach in weakly supervised text classification~\cite{DBLP:conf/cikm/MengSZ018, mekala-shang-2020-contextualized, DBLP:conf/emnlp/MengZHXJZH20, wang-etal-2021-x} and train a final text classifier based on the pseudo-labels assigned to each text. This usually improves the performance as the fine-tuned classifier mitigates some noisy in the pseudo-labels.
\section{Experiments}

\subsection{Datasets}
We evaluate \our on 7 popular datasets of different textual sources and criteria of classes, including news article datasets 20News~\cite{lang1995newsweeder}, NYT-Small~\cite{DBLP:conf/cikm/MengSZ018}, NYT-Topics~\cite{10.1145/3366423.3380278}, NYT-Locations~\cite{10.1145/3366423.3380278} and AGNews~\cite{zhang2015character}, an internet question-answering dataset Yahoo~\cite{NIPS2015_250cf8b5}, and an ontology categorization dataset DBpedia~\cite{zhang2015character} based on 14 ontology classes in DBpedia.
Table~\ref{tab:0} contains the detailed statistics of the 7 datasets.



Sentiment analysis is also popular in text classification. However, many explored sentiment analysis settings with weak supervision are on the coarse-grained setting~\cite{wang-etal-2021-x,DBLP:conf/emnlp/MengZHXJZH20} with too less classes (e.g., positive and negative), which is not practical for open-world class detection.

\subsection{Compared Methods}
We compare our method with 3 open-world classification methods, Rankstats+, ORCA and GCD.

\textbf{Rankstats}~\cite{han2021autonovel} (aka AutoNovel) is the first approach crafted for open-world classification without relying on any human annotation of the unseen classes. It tackles the task by joint learning to transfer knowledge from labeled to unlabeled data with ranking statistics.
Since its original setting requires the labeled and unlabeled classes to be disjoint, we follow the process in GCD paper~\cite{vaze2022gcd} to adapt it to our setting and named it \textbf{Rankstats+}.
\textbf{ORCA}~\cite{cao2021openworld} is a general method for open-world semi-supervised classification, which further reduces the supervision of the seen classes. It utilizes an uncertainty adaptive margin to reduce the learning gap between seen and unseen classes.
\textbf{GCD}~\cite{vaze2022gcd} is also semi-supervised, utilizing contrastive representation learning and clustering to  directly provide class labels, and improve Rankstats' method of estimating the number of unseen classes.
To adapt these three methods to the text domain, we use BERT on top of their frameworks to obtain the document representations as the training feature. And since these three methods utilize self-supervised learning grounded upon visual data, within the text domain, we harness the SimCSE~\cite{gao2021simcse} approach.

We also propose other two baselines.
BERT is known to capture the domain information of a document well~\cite{aharoni2020unsupervised, DBLP:conf/emnlp/WangDS21}.
So we design \textbf{BERT+GMM}, which utilizes the CLS token representations after fine-tuning on the labeled dataset to fit a Gaussian Mixture Model for all classes. We also propose \textbf{BERT+SVM}. We first utilize a Support Vector Machine~\cite{hearst1998support} to find all outlier documents based on CLS token representations and then classify documents belonging to seen classes with a BERT classifier and cluster outlier documents.

\begin{table*}[t!]
  \begin{center}
    \caption{Evaluations of compared methods and \our. The overall mean micro-/macro-F$_1$ scores over three runs are reported. We also report performances for seen and unseen classes separately in Table \ref{table:app}, \ref{table:app_}.
    }\label{table:1}
    \resizebox{\linewidth}{!}{
    \begin{tabular}{c  c  c  c c c  c c c c}
      \toprule
      \textbf{Method} & \textbf{Extra Info} & \textbf{AGNews} & \textbf{20News} & \textbf{NYT-S} & \textbf{NYT-Top} & \textbf{NYT-Loc} & \textbf{Yahoo} & \textbf{DBpedia} & \textbf{Average}\\
       \midrule
       Rankstats+ & \multirow{4}{*}{\large \usym{2717}} & 39.53/28.55 & 24.94/13.88 & 52.01/23.13 & 42.23/19.98 & 39.68/23.13 & 29.66/20.44 & 48.20/39.15 & 39.47/24.04\\
       ORCA & & 72.44/72.27 & 48.92/39.83 & 74.34/42.22 & 62.23/39.02 & 58.71/44.81 & 35.57/32.71 & 69.27/67.92 & 60.21/48.40\\
       GCD & & 66.37/66.51 & 51.75/42.96 & 82.59/63.35 & 66.36/39.69 & 70.25/53.41 & 36.73/35.39 & 75.81/72.97 & 64.27/53.47\\
      \our & & \textbf{79.42}/\textbf{79.75} & \textbf{79.07}/\textbf{79.29} & \textbf{94.78}/\textbf{88.46} & \textbf{78.67}/\textbf{69.48} & \textbf{80.94}/\textbf{79.55} & \textbf{54.46}/\textbf{56.23} & \textbf{85.15}/\textbf{84.87} & \textbf{78.93}/\textbf{76.80}\\
      \midrule
      Rankstats+ (OE)&  \multirow{5}{*}{\# of Classes} & 61.44/57.50 & 53.65/38.12 & 40.82/31.67 & 19.93/15.07 & 21.96/16.81 & 32.79/26.94 & 50.03/44.31 & 40.09/32.92\\
      ORCA (OE) & & 64.38/64.50 & 51.85/40.04 & 70.44/46.21 & 59.42/38.29 & 42.99/33.08 & 43.87/41.43 & 82.54/81.30 & 59.35/49.26\\
      GCD (OE) & & 65.42/65.44 & 61.27/56.42 & 78.82/56.59 & 70.51/42.44 & 55.37/44.86 & 39.01/37.58 & 84.14/83.60 & 64.93/55.28\\
      BERT+GMM (OE) & & 38.25/37.14 & 29.32/25.21 & 58.79/24.79 & 26.88/14.08 & 11.64/9.47 & 14.11/13.64 & 14.74/14.20 & 27.68/19.79\\
      BERT+SVM (OE) & & 45.20/44.15 & 39.07/34.96 & 51.97/22.34 & 24.95/12.83 & 13.91/7.45 & 15.25/13.39 & 16.28/14.41 & 29.52/21.36\\
      \bottomrule
    \end{tabular}}
  \end{center}
\end{table*}

\begin{table*}[t!]
  \begin{center}
    \caption{Performance of compared methods and \our on different imbalance degrees.
    We report macro-F$_1$ scores to more effectively demonstrate the results under an imbalanced data distribution.
    For all compared methods, we report their Pareto optimal started with 100 classes and our estimation. 
    }\label{tab:per-im}
    \begin{tabular}{c  c c c  c c c  c c c  c c c}
      \toprule
      \multirow{2}{*}{\textbf{Method}} &
      \multicolumn{3}{c}{\textbf{DBpedia}} & \multicolumn{3}{c}{\textbf{DBpedia-Low}} & \multicolumn{3}{c}{\textbf{DBpedia-Medium}} & \multicolumn{3}{c}{\textbf{DBpedia-High}}\\
       & \textbf{All} & \textbf{Seen} & \textbf{Unseen} & \textbf{All} & \textbf{Seen} & \textbf{Unseen} & \textbf{All} & \textbf{Seen} & \textbf{Unseen} & \textbf{All} & \textbf{Seen} & \textbf{Unseen}\\
      \midrule
      ORCA & 81.30 & 95.19 & 67.42 & 76.07 & 95.51 & 56.64 & 72.21 & 97.20 & 47.23 & 69.99 & 97.63 & 42.34 \\
      GCD & 83.60 & 93.48 & 73.71 & 82.92 & 94.32 & 71.51 & 81.78 & 94.42 & 69.15 & 75.57 & 92.53 & 58.61 \\
      \our & 84.87 & 87.16 & 82.59 & 85.81 & 91.82 & 79.97 & 84.97 & 93.31 & 76.63 & 79.35 & 88.81 & 69.90 \\
      \bottomrule
    \end{tabular}
  \end{center}
\end{table*}

\begin{table}[t!]
  \begin{center}
  \scriptsize
    \caption{An overview of the imbalanced DBpedia datasets with 14 classes. 
    }\label{table:imbalance}
    \resizebox{0.85\linewidth}{!}{
    \begin{tabular}{c  c  c  c}
      \toprule
        & \textbf{Low} & \textbf{Medium} & \textbf{High}\\
      \midrule
      $\Delta$ & 2\% & 4\% & 6\%\\
      \# of Documents & 19,480  & 16,565 & 13,652 \\
      Imbalance & 1.35 & 2.09 & 4.56\\
      \bottomrule
    \end{tabular}}
  \end{center}
\end{table}

\begin{figure}[t!]
\vspace{-0.1cm}
\begin{center}
\includegraphics[width = .35\textwidth]{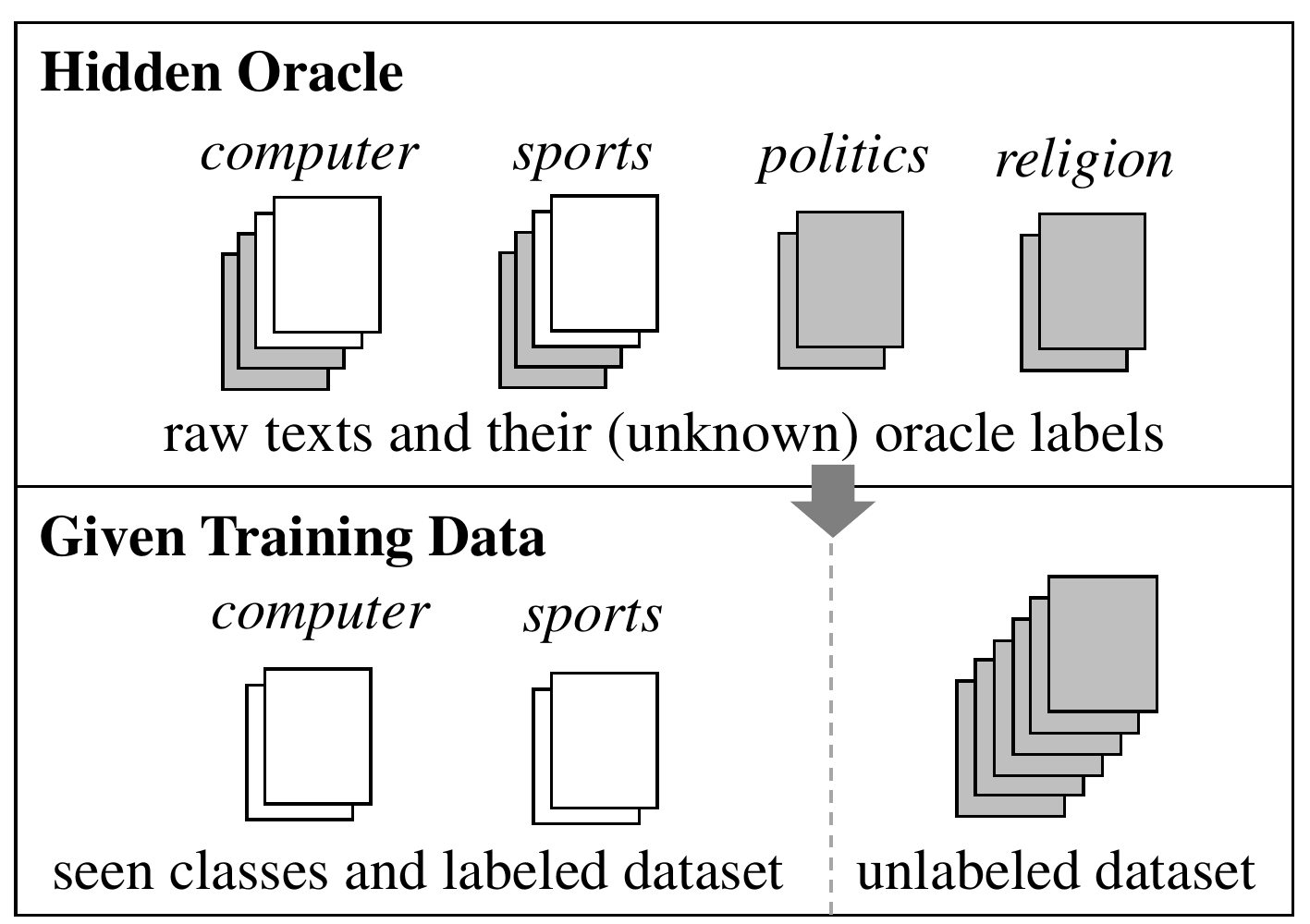} 
\caption{Schematic diagram of the corpus split. Only 10 samples in popular classes are provided as training labels.
}
\label{fig:setting}
\end{center}
\vspace{-0.3cm}
\end{figure}

\subsection{Experimental Settings}~\label{sec:4.3}
For the basic experiments, we split the classes into half seen and half unseen. We set the most infrequent half of classes (i.e., classes with fewer documents) as unseen, which emphasizes the imbalanced and emerging nature of real-world scenarios. Among the seen classes, we give 10-shot supervision, that is 10 documents for each seen class containing labels and the rest are unlabeled (Figure~\ref{fig:setting}).

For \our and all compared methods in our experiments, we utilize the pre-trained bert-base-uncased model provided in Huggingface’s Transformers library~\cite{wolf2019huggingface}.
All experiments are conducted using a 32-core processor and a single NVIDIA RTX A6000 GPU.
For \our, the default hyper-parameter settings are $K=100$, $W=50$, and $\beta=0.7$. The analysis of hyper-parameter sensitivity is presented in Section~\ref{sec:4.4}.

Since all compared methods require the total number of classes as input, we evaluate them in two ways.
\begin{itemize}[leftmargin=*,nosep]
\item \textbf{Our Estimation (OE)}: 
To ensure the comparison is fair to the baseline methods, we also run all the baselines based on (3 runs) of our prediction of classes.

\item \textbf{Baselines' Estimation}:
Since Rankstats+, ORCA, and GCD can also work started with an overestimated number of the initial classes and provide the prediction of the classes. 
So we also test these three methods starting with $K=100$ classes as same as our method. Since further experiments show their predictions don't work well in most of our datasets, we do not test other baselines with their estimations.
\end{itemize}

\begin{table*}[t!]
  \begin{center}
    \caption{Performance of seen classes. The mean micro/macro-F$_1$ scores over three runs are reported.}\label{table:app}
    \resizebox{\linewidth}{!}{
    \begin{tabular}{c  c  c  c c c  c c c c}
      \toprule
      \textbf{Method} & \textbf{Extra Info} & \textbf{AGNews} & \textbf{20News} & \textbf{NYT-S} & \textbf{NYT-Top} & \textbf{NYT-Loc} & \textbf{Yahoo} & \textbf{DBpedia} & \textbf{Average}\\
       \midrule
       Rankstats+ & \multirow{4}{*}{\large \usym{2717}} & 52.74/57.10 & 33.35/34.71 & 58.17/51.15 & 46.75/44.73 & 44.28/46.26 & 39.59/40.88 & 68.29/71.87 & 49.02/49.53\\
       ORCA & & 75.34/75.29 & 64.57/59.41 & 85.90/77.45 & 72.14/67.07 & 70.96/71.92 & 43.92/40.73 & \textbf{97.99}/\textbf{98.00} & 72.97/69.98\\
       GCD & & 66.47/66.59 & 66.80/67.97 & 90.93/82.15 & 78.56/72.09 & 82.00/79.78 & 48.04/48.83 & 91.36/92.23 & 74.88/72.81\\
      \our & & \textbf{77.72}/\textbf{77.76} & \textbf{85.60}/\textbf{85.86} & \textbf{97.33}/\textbf{91.68} & \textbf{83.04}/\textbf{81.92} & \textbf{84.13}/\textbf{88.10} & \textbf{52.50}/\textbf{56.14} & 87.69/87.16 & \textbf{81.14}/\textbf{81.23}\\
      \midrule
      Rankstats+ (OE) & \multirow{5}{*}{\# of Classes} & 57.93/52.83 & 53.33/50.40 & 50.82/52.64 & 24.75/29.01 & 26.84/27.03 & 43.14/41.92 & 76.49/81.64 & 47.62/47.92\\
      ORCA (OE) & & 74.15/74.15 & 68.60/69.71 & 84.93/78.52 & 71.19/65.74 & 55.70/54.12 & 53.13/36.77 & 95.18/95.19 & 71.84/67.74\\
      GCD (OE) & & 65.36/65.35 & 75.07/73.60 & 89.32/80.33 & 82.82/73.01 & 70.58/69.25 & 49.83/49.40 & 93.13/93.48 & 75.16/72.06\\
      BERT+GMM (OE) & & 31.67/31.01 & 36.14/34.77 & 69.11/41.76 & 34.54/23.84 & 13.73/12.18 & 14.57/14.17 & 16.28/15.70 & 30.86/24.78\\
      BERT+SVM (OE) & & 40.40/37.96 & 47.29/45.75 & 55.29/39.87 & 29.49/20.48 & 17.13/8.79 & 15.74/12.93 & 14.61/11.61 & 31.42/25.34\\
      \bottomrule
    \end{tabular}}
  \end{center}
\end{table*}

\begin{table*}[t!]
  \begin{center}
    \caption{Performance of unseen classes. The mean micro/macro-F$_1$ scores over three runs are reported.}\label{table:app_}
    \resizebox{\linewidth}{!}{
    \begin{tabular}{c  c  c  c c c  c c c c}
      \toprule
      \textbf{Method} & \textbf{Extra Info} & \textbf{AGNews} & \textbf{20News} & \textbf{NYT-S} & \textbf{NYT-Top} & \textbf{NYT-Loc} & \textbf{Yahoo} & \textbf{DBpedia} & \textbf{Average}\\
       \midrule
       Rankstats+ & \multirow{4}{*}{\large \usym{2717}} & 0/0 & 0/0 & 13.35/13.35 & 0/0 & 0/0 & 0/0 & 18.69/8.49 & 4.58/3.12\\
       ORCA & & 69.19/69.26 & 27.59/26.78 & 23.96/18.73 & 22.86/16.58 & 19.27/17.70 & 25.03/24.70 & 40.65/37.85 & 32.65/30.23\\
       GCD & & 66.26/66.43 & 32.98/28.42 & 55.29/50.82 & 17.33/15.32 & 31.20/27.03 & 24.84/22.88 & 62.28/58.74 & 41.45/38.52\\
      \our & & \textbf{81.29}/\textbf{82.22} & \textbf{73.65}/\textbf{74.91} & \textbf{86.45}/\textbf{86.31} & \textbf{62.07}/\textbf{59.52} & \textbf{72.45}/\textbf{70.99} & \textbf{57.00}/\textbf{56.34} & \textbf{82.51}/\textbf{82.59} & \textbf{73.63}/\textbf{73.27}\\
      \midrule
      Rankstats+ (OE) & \multirow{5}{*}{\# of Classes} & 64.44/62.17 & 35.56/29.93 & 16.17/17.68 & 8.48/3.92 & 7.60/6.59 & 23.07/11.97 & 18.13/6.97 & 24.78/19.89\\
      ORCA (OE) & & 55.26/54.84 & 24.70/20.27 & 26.47/24.66 & 20.34/16.32 & 15.58/12.04 & 34.43/30.70 & 69.74/67.42 & 35.22/32.32\\
      GCD (OE) & & 65.45/65.52 & 47.33/44.97 & 40.66/40.75 & 19.83/17.98 & 19.98/20.47 & 27.20/25.76 & 74.64/73.71 & 42.15/41.31\\
      BERT+GMM (OE) & & 43.74/43.28 & 20.03/18.84 & 14.12/13.46 & 6.80/6.28 & 7.54/6.76 & 13.52/13.10 & 12.96/12.70 & 16.96/16.35\\
      BERT+SVM (OE) & & 51.22/50.33 & 27.11/27.76 & 10.90/10.89 & 7.33/6.83 & 6.52/6.11 & 14.44/13.85 & 17.67/17.21 & 19.31/19.00\\
      \bottomrule
    \end{tabular}}
  \end{center}
\end{table*}

\smallsection{Evaluation}
There are several evaluation criteria like accuracy, or NMI-based clustering metrics have been used in previous work~\cite{han2021autonovel, cao2021openworld, vaze2022gcd}. However, they were proposed in a balanced setting and would be biased toward the popular classes when the classes are imbalanced (i.e., the low penalty for misclassification of infrequent unseen classes). Since we argue that in open-world classification, the new, emerging classes are naturally the minority classes, these metrics are not suitable. 

Therefore, we propose a new evaluation criterion based on F$_1$ Score to better demonstrate results. Since the final number of classes produced by a method may not equal the ground truth, a mapping from the prediction to the actual classes is required. 
Given the clusters of documents provided by a method and the ground-truth classes of documents, we first perform a maximum bipartite matching between the method-provided clusters and the ground-truth classes, where the edge weights are the number of overlapping documents between the clusters and the ground-truth classes.\footnote{In the case when the number of predicted clusters is less than the ground truth, we create virtual clusters with no documents inside.} 
The matched clusters are assigned to the corresponding classes. This step is to guarantee that all classes have some predictions.
For each remaining cluster, we simply assign it to the class with which it exhibits the maximal matches. 

This is the equation. Consider a matrix $\mathcal M$, where $\mathcal M_{i, j}$ denotes the number of text in cluster $i$ that belongs to class $j$. We use $r_i$ to denote the assignment of each cluster:

\begin{equation*}\footnotesize
r_i = 
    \begin{cases}
        j,  & \text{where $i$, $j$ are obtained by maximum matching on }\mathcal M; \\
        \arg\max_j(\mathcal M_{i, j}), & \text{otherwise.} \\
    \end{cases}
\end{equation*}

After assigning all clusters to the classes, the F$_1$ score can be computed instance-wise on the text as the evaluation criterion for classification performance. 

\subsection{Experimental Results}\label{sec:4.4}

\smallsection{\our Performance} We assess the weakly supervised open-world performance of \our versus other baselines. Table~\ref{table:1} contains overall comparisons, Table~\ref{table:app} and \ref{table:app_} further provide performances on seen and unseen classes.
Specifically, \our outperforms BERT+GMM and BERT+SVM across all 7 datasets for both seen and unseen classes, even though they are given the matching number of classes as input. This strengthens the need for our iterative refinement process since merely applying few-shot tuning does not bring as good performance as ours.
Moreover, \our performs noticeably better than the general methods Rankstats+, ORCA, and GCD under the same circumstances.
Even when the matching number is given as input to them, \our consistently outperforms them in all cases on all datasets, except for 
the seen part of DBpedia.
\begin{figure}[t!]
\begin{center}
\includegraphics[width = .48\textwidth]{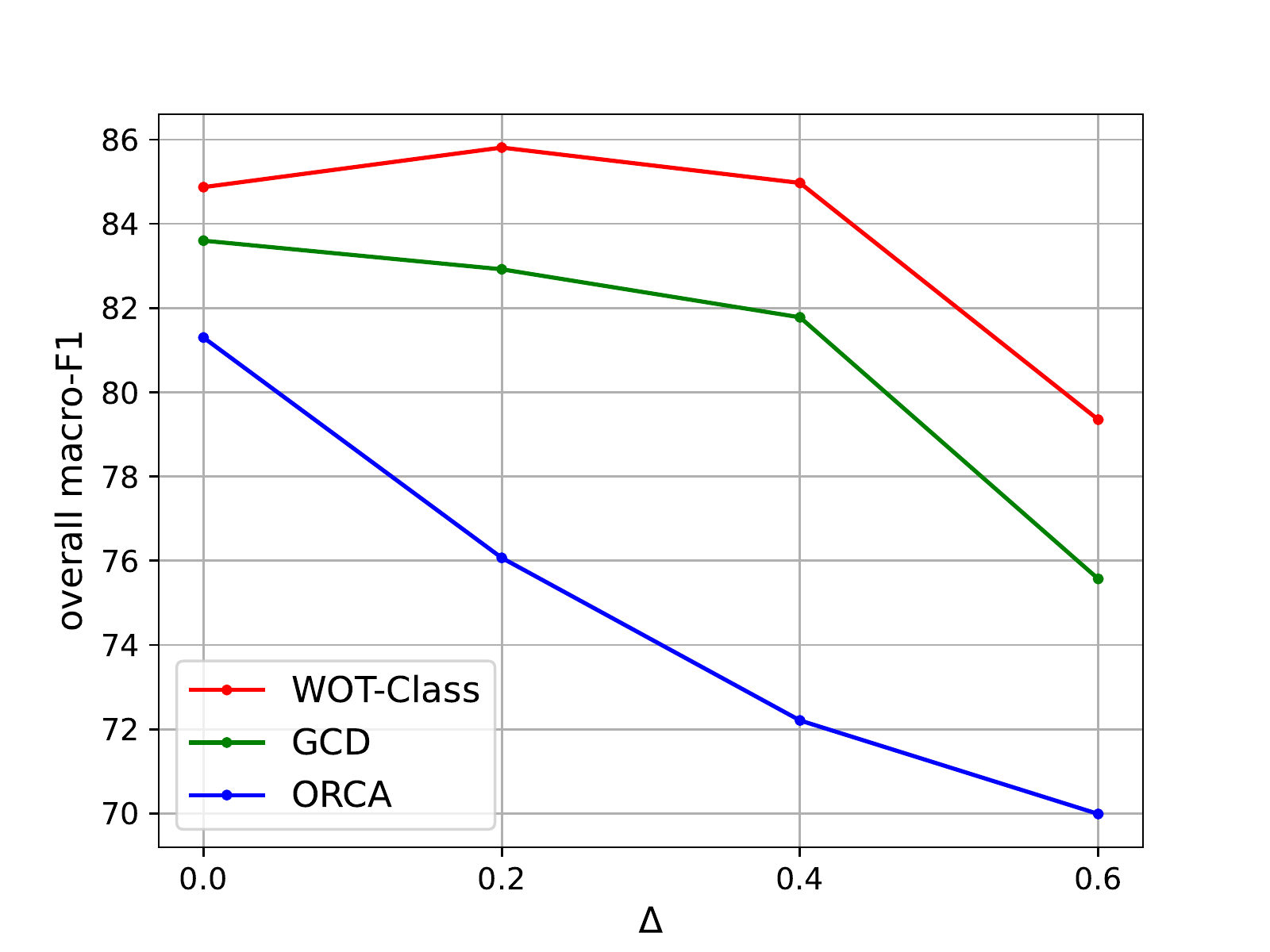} 
\caption{Overall performance of compared methods and \our on different imbalance degrees.}
\label{fig:imb}
\end{center}
\end{figure}

\begin{table*}[t!]
  \begin{center}
    \caption{Predictions of the number of classes. The average and standard deviation over three runs are reported. The average offset refers to the average of the absolute discrepancies between each prediction and the ground truth value.}~\label{table:prediction}
    \resizebox{\linewidth}{!}{
    \begin{tabular}{c  c  c  c  c  c  c  c  c}
      \toprule
      \textbf{Method} & \textbf{AGNews} & \textbf{20News} & \textbf{NYT-S} & \textbf{NYT-Top} & \textbf{NYT-Loc} & \textbf{Yahoo} & \textbf{DBpedia} & \textbf{Average Offset}\\
      \midrule
      Rankstats+ & 2.00$_0$ & 2.00$_0$ & 2.33$_{0.58}$ & 4.33$_{0.58}$ & 5.00$_0$ & 5.00$_0$ & 10.33$_{3.06}$ & 3.71 \\
      ORCA & 69.00$_{6.08}$ & 53.67$_{45.65}$ & 39.33$_{32.35}$ & 96.33$_{2.89}$ & 94.67$_{3.21}$ & 66.67$_{55.16}$ & 66.00$_{3.61}$ & 62.48\\
      GCD & 23.33$_{2.89}$ & 16.00$_{19.92}$ & 59.33$_{17.01}$ & 29.67$_{26.63}$ & 27.33$_{9.24}$ & 20.00$_{16.64}$ & 14.00$_{3.46}$ & 19.81\\
      \our & 19.67$_{1.15}$ & 20.67$_{0.58}$ & 27.00$_{2.89}$ & 18.67$_{3.61}$ & 11.67$_{0.58}$ & 21.00$_{2.00}$ & 17.67$_{2.89}$ & 11.33\\
      Ground Truth & 4 & 5 & 5 & 9 & 10 & 10 & 14 & -\\
      \bottomrule
    \end{tabular}}
  \end{center}
\end{table*}

\begin{figure*}[t!]
\begin{center}
\subfigure[$K$]{
    \includegraphics[width =.32\linewidth]{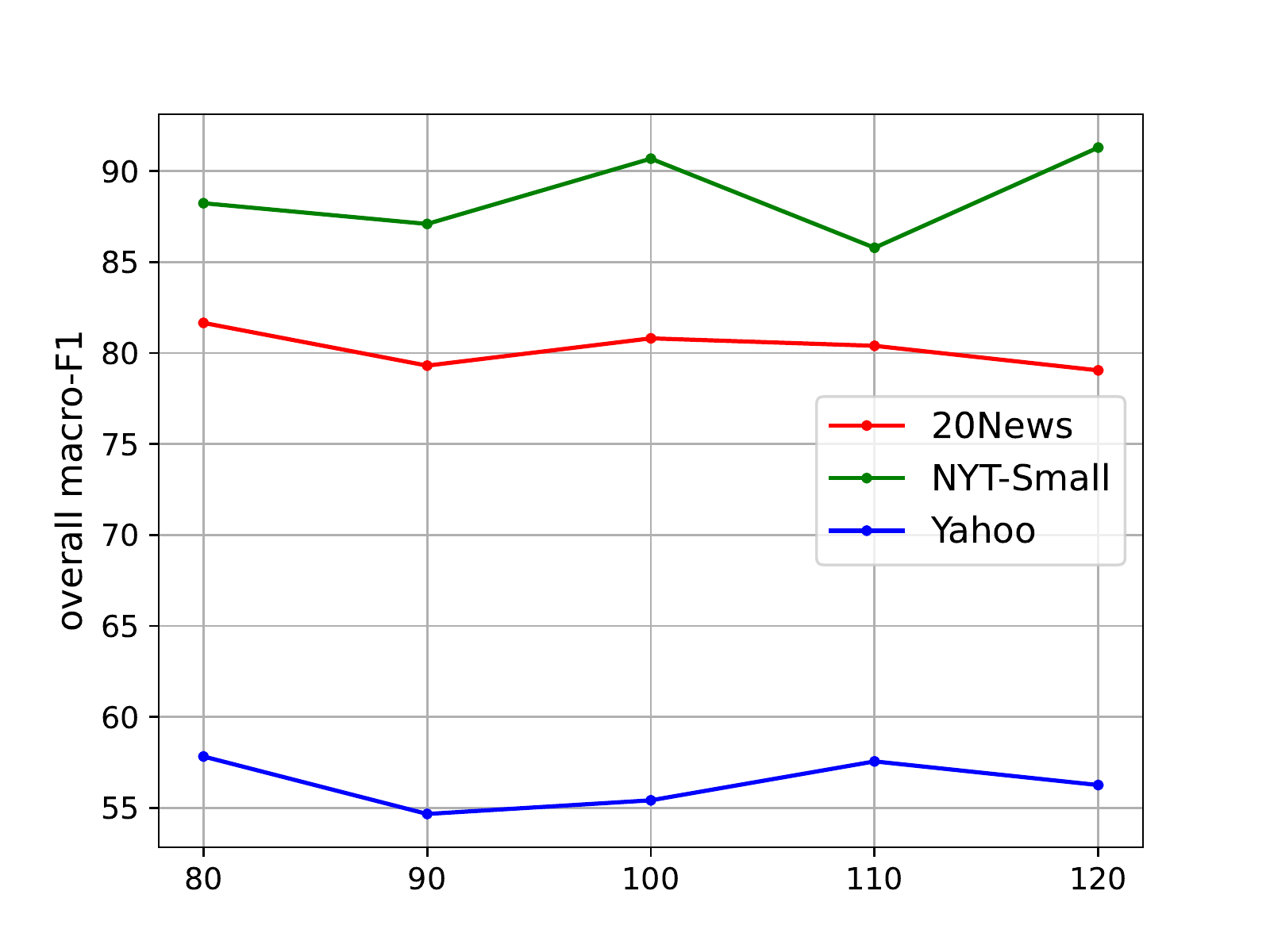}
}
\subfigure[$W$]{
    \includegraphics[width =.32\linewidth]{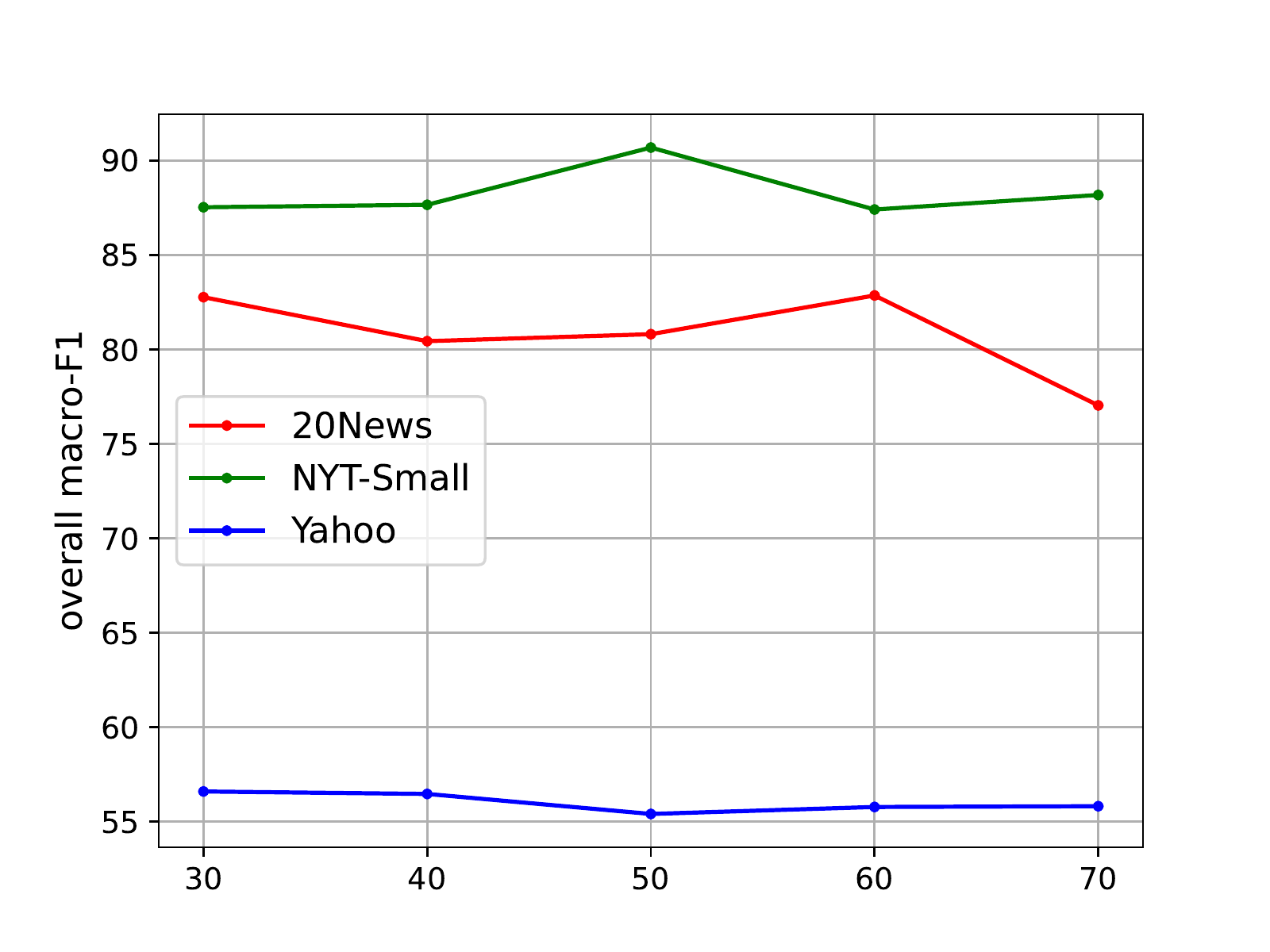}
}
\subfigure[$\beta$]{
    \includegraphics[width =.32\linewidth]{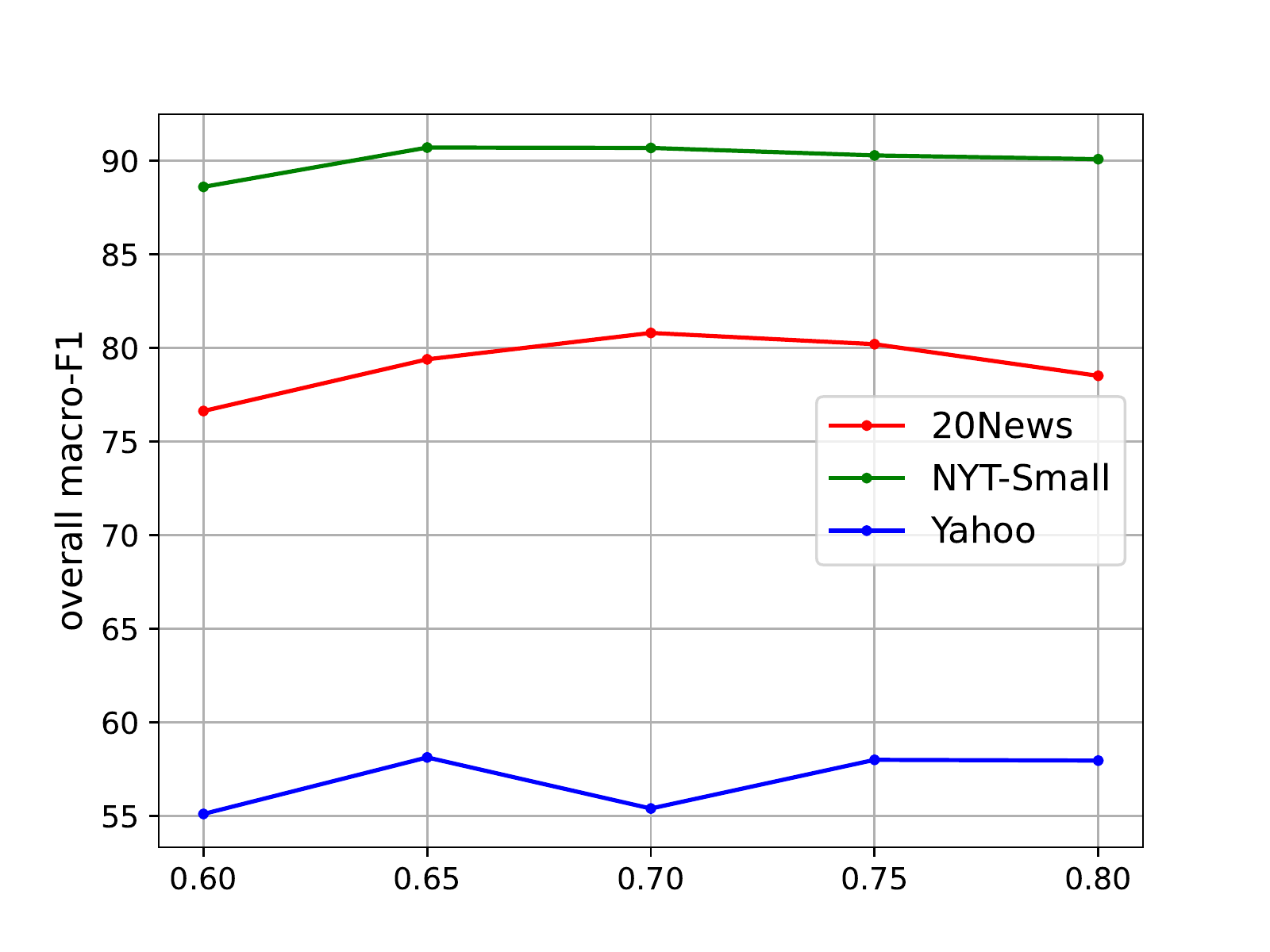}
}
\caption{Hyper-parameter sensitivity study on 20News, NYT-Small and Yahoo. The overall macro-F$_1$ scores using a fixed random seed are reported.}
\label{Fig.hyper}
\end{center}
\end{figure*}

\begin{table}[t!]
  \begin{center}
    \caption{Examples of the class-words. We use '[]' to split class-words belonging to different clusters in \our.}\label{table:4}
     \resizebox{0.95\linewidth}{!}{
    \begin{tabular}{ c  c  c }
      \toprule
        \textbf{Dataset} & \textbf{Ground Truth} & \textbf{\our} \\ \midrule
        \multirow{5}{*}{NYT-Loc} & \texttt{Russia} & \texttt{[Ukraine, Russia]}\\
        & \texttt{Germany} & \texttt{[Germany]}\\
        & \texttt{Canada} & \texttt{[Canada]}\\
        & \texttt{France} & \texttt{[France]}\\
        & \texttt{Italy} & \texttt{[Italy]}\\
        \midrule
        \multirow{8}{*}{DBpedia} & \texttt{athlete} & \texttt{[footballer], [Olympics]}\\ 
		& \multirow{2}{*}{\texttt{artist}} & \texttt{[painting, painter, art],}\\
        & & \texttt{[tv, theatre, television]}\\ 
        & \texttt{company} & \texttt{[retail, company, business]}\\ 
        & \texttt{school} & \texttt{[school, education, academic]}\\ 
        & \texttt{politics} & \texttt{[politician]}\\ 
        & \texttt{transportation} & \texttt{[aircraft, locomotive]}\\ 
        & \texttt{building} & \texttt{[architecture, tower, church]}\\
        \bottomrule
    \end{tabular}
    }
  \end{center}
\end{table}

\smallsection{Imbalance Tolerance}
As generic solutions, ORCA, and GCD with our prediction number of classes only have a little performance margin with \our on balanced dataset DBpedia, while underperforming more severely on other imbalanced datasets. To gain further insights, we conduct experiments on the tolerance of imbalance for \our and these three compared methods.

As shown in Table~\ref{table:imbalance}, we construct three imbalanced DBpedia datasets with different degrees of imbalance. This is achieved by removing the number of samples in each class by a linearly increasing ratio $\Delta$. For example, when $\Delta = 4\%$, the classes have $100\%$, $96\%$, $92\%, \ldots$ of its original documents. We choose the ordering of classes randomly but fixed across the Low, Medium, and High experiments, and by design, the classes with a larger number of documents are seen classes.

Figure~\ref{fig:imb} and Table~\ref{tab:per-im} shows the result of \our and compared methods on the constructed DBpedia. ORCA and GCD are sensitive to imbalanced classes, especially for the unseen part of the data. Even after reporting the Pareto optimal results based on their own predictions and our estimation for these two methods, their overall performance still dropped by 11.31\% and 8.03\% respectively as the data distribution became more imbalanced, while our method experienced a relative drop of only 5.52\%.  
This experiment shows that \our is more robust to imbalanced classes of text datasets which are common in the real world (e.g., the imbalance ratio of NYT collected from NYT News is 16.65).

\smallsection{Prediction of the Number of Classes}\label{sec:ClassNum}
\our starts with an initial guess on the number of classes and removes redundant ones iteratively.
The number of the remaining classes is its prediction of the total number of classes.
As shown in Table~\ref{table:prediction}, in most cases, \our's final predicted number of classes is around 2 to 4 times larger than the ground truth, which is affordable for human inspection.
And the estimation turns out to be reasonable as shown in Table~\ref{table:4}, \our overestimates because its predicted classes are the fine-grained version of the ground truth classes. For example, DBpedia's \textit{artist} class can be split into two classes which respectively related to painting and performing. 
Moreover, our class-words are highly related to (or even as same as) ground truth class names and human-understandable.
So based on our prediction of classes with class-words, users can simply find some underlying sub-classes and decide whether to merge them.

As baselines, Rankstats+, ORCA, and GCD can also start with an overestimated number of classes and get a prediction of classes.
However, given the same initial number of classes, Rankstats+ struggles to discover any classes beyond the seen classes, and ORCA hardly eliminates enough classes to provide the user with an intuitive understanding of the dataset's text distribution. 
GCD can provide more reasonable predictions, but compared to our approach, its prediction still deviates substantially from the ground truth and is much more unstable. 
This indicates these methods' ability to estimate the number of classes in the few-shot setting is not reliable.

\smallsection{Hyper-parameter Sensitivity}
\our has 3 hyper-parameters: $K$, $W$, $\beta$, and we show their default values in Sec.~\ref{sec:4.3}.
To further explore the stability and robustness of \our, we conduct a hyper-parameter sensitivity study on three datasets with different imbalance rates: 20News, NYT-Small and Yahoo, to study how fluctuations in hyper-parameters influence the performance of our method. The experiment is performed using a fixed random seed (42) for reproducibility. We report the overall macro-F$_1$ scores for 5 distinct values of each hyperparameter. 
As illustrated in Figure~\ref{Fig.hyper}, the performance fluctuations remain within reasonable margins, basically under 5\%. Our method does not need to fine-tune these hyper-parameters.

\section{Related Work}

\smallsection{Open-world Learning}
Traditional open-world recognition methods~\cite{bendale2015towards, rudd2017extreme, boult2019learning} aim to incrementally extend the set of seen classes with new unseen classes. These methods require human involvement to label new classes.

Recently, Rankstats~\cite{han2021autonovel} first defined open-world classification tasks with no supervision on unseen classes in the image domain and proposed a method with three stages. The model begins with self-supervised pre-training on all data to learn low-level representations. It is then trained under full supervision on labeled data to glean higher-level semantic insights. Finally, joint learning with ranking statistics is performed to transfer knowledge from labeled to unlabeled data.
However, Rankstats still require full supervision on seen classes to get high performance.

Following that, ORCA~\cite{cao2021openworld} and GCD~\cite{vaze2022gcd} defined open-world semi-supervised classification and proposed general solutions which further improved the framework of Rankstats and alleviated the burden of manual annotation. 
However, these methods' performance is not robust enough for the few-shot task and the imbalanced data distribution in the text domain.
In contrast, our work is applicable to infrequent classes 
and exploits the fact that the input is words which are class-indicative.

\smallsection{Extremely Weak Supervision in NLP}
\citet{aharoni2020unsupervised} showed that the average of BERT token representations can preserve documents' domain information.
X-Class~\cite{wang-etal-2021-x} followed this idea to propose the extremely weak supervision setting where text classification only relies on the name of the class as supervision.
However, such methods can not transfer to open-world classification naively as they cannot detect unseen classes.
Our method leverages such extremely weak supervision methods as a subroutine to help the clustering of documents. 
But importantly, we note that such methods cannot be applied straightforwardly as they also are sensitive to noise and too similar classes. 
We show that our general idea of using class-words can further help an extremely weak supervision method to obtain stable performance.

\smallsection{Joint Clustering with Downstream Tasks}
To some sense, our method leverages partly an idea called joint clustering, which some recent works~\cite{caron2018deep,asano2020self} in the image domain achieved high performance through jointly performing clustering and image classification.
Their main idea is to utilize clustering to extract the hidden information of image representations and generate pseudo-labels, which in turn provide supervision for classification training and ultimately guide the co-improvement of representation and clustering.
However, the crucial difference is that their methods already know the predefined classes and highly depend on strong assumptions like all classes share the same size to obtain excellent performance.
Conversely, \our utilizes the general idea of joint clustering in an open-world setting where the classes may be too fine-grained and noisy. We address these unique challenges via the class-words we propose and show that our methodology can not only estimate the precise number of classes but also tolerate imbalanced data distribution.
\section{Conclusions and Future Work}
In this paper, we introduce the challenging yet promising weakly supervised open-world text classification task. We have identified the key challenges and unique opportunities of this task and proposed \our that achieves quite decent performance with minimal human effort. 
Specifically, \our starts with an overestimated number of classes and constructs an iterative refinement framework that jointly performs the class-word ranking and document clustering, leading to iterative mutual enhancement.
Consequently, \our can progressively extract the most informative classes and assemble similar documents, resulting in an effective and stable open-world classification system, which is validated by comprehensive experiments.
In the future, we envision that open-world text classification can be conducted with even less manual annotation, for example, by only requiring user-provided hyper-concept (e.g., Topics, Locations) or custom instructions. This will further reduce the cost of classification systems and extend their applicability.

In summary, this paper presents an initial exploration of open-world text classification, including problem formulation, methodology, and empirical results. We hope this work can inspire more research in open-world learning for NLP. As an emerging field, open-world classification demands more algorithms, datasets, and evaluation metrics to truly unleash its potential.


\bibliographystyle{ACM-Reference-Format}
\balance
\bibliography{sample-base}


\end{document}